\documentclass[letterpaper]{article} 
\usepackage{aaai2026}  
\usepackage{times}  
\usepackage{helvet}  
\usepackage{courier}  
\usepackage[hyphens]{url}  
\usepackage{graphicx} 
\usepackage{natbib}  
\usepackage{caption} 
\usepackage{algorithm}

\usepackage{newfloat}
\usepackage{listings}
\DeclareCaptionStyle{ruled}{labelfont=normalfont,labelsep=colon,strut=off} 
\floatstyle{ruled}
\newfloat{listing}{tb}{lst}{}
\floatname{listing}{Listing}

\usepackage[table]{xcolor}

\usepackage{amsmath,amsfonts}
\usepackage{array}
\usepackage{textcomp}
\usepackage{url}
\usepackage{verbatim}
\usepackage{graphicx}

\usepackage{amsmath}
\usepackage{booktabs}
\usepackage{multirow}
\usepackage[table]{xcolor} 

\newcommand{\ie}{\textit{i.e.} }
\newcommand{\eg}{\textit{e.g.}}

\usepackage[utf8]{inputenc} 
\usepackage[T1]{fontenc}    

\usepackage{url}            
\usepackage{booktabs}       
\usepackage{amsfonts}       
\usepackage{nicefrac}       
\usepackage{microtype}      
\usepackage{xcolor}         

\usepackage{epsfig}
\usepackage{graphicx}
\usepackage{amsmath}
\usepackage{amssymb}
\usepackage{multirow} 
\usepackage{caption} 

\usepackage{cleveref}
\crefname{figure}{Fig.}{Figs.}
\crefname{table}{Tab.}{Tabs}

\usepackage{booktabs}
\usepackage{multirow}
\usepackage[table]{xcolor} 

\usepackage{algorithm}
\usepackage{algpseudocode}

\usepackage{etoolbox}
\AtBeginEnvironment{algorithmic}{%
  \setlength{\baselineskip}{1.2\baselineskip}
  \setlength{\itemsep}{3pt}
  \setlength{\parskip}{2pt}
}

\usepackage{amsthm}        
\usepackage{amsmath}          
\usepackage{graphicx}      
\usepackage{booktabs}      
\usepackage{xcolor}

\usepackage{booktabs}
\usepackage{multirow}
\usepackage{makecell}
\usepackage{array}

\title{Creating Blank Canvas Against AI-enabled Image Forgery}

\author{
    Qi Song, 
    Ziyuan Luo, 
    Renjie Wan\thanks{Corresponding author. }
}
\affiliations{
    Department of Computer Science, Hong Kong Baptist University\\
    \texttt{
    \{qisong, ziyuanluo\}@life.hkbu.edu.hk, 
    \{renjiewan\}@hkbu.edu.hk}
}

\usepackage{bibentry}

\begin{document}

\maketitle

\begin{abstract}
AIGC-based image editing technology has greatly simplified the realistic-level image modification, causing serious potential risks of image forgery. This paper introduces a new approach to tampering detection using the Segment Anything Model (SAM). Instead of training SAM to identify tampered areas, we propose a novel strategy.  The entire image is transformed into a blank canvas from the perspective of neural models.  Any modifications to this blank canvas would be noticeable to the models. To achieve this idea, we introduce adversarial perturbations to prevent SAM from ``seeing anything'', allowing it to identify forged regions when the image is tampered with. Due to SAM's powerful perceiving capabilities, naive adversarial attacks cannot completely tame SAM. To thoroughly deceive SAM and make it blind to the image, we introduce a frequency-aware optimization strategy, which further enhances the capability of tamper localization. Extensive experimental results demonstrate the effectiveness of our method.
\end{abstract}   

\begin{links}
    \link{Code}{https://github.com/qsong2001/blank_canvas}
\end{links}

\section{Introduction}

Image tamper localization aims to mitigate the proliferation of forged imagery that threatens public trust and social stability. Contemporary AIGC frameworks~\cite{rombach2022high, podell2023sdxl, zhang2023adding, lugmayr2022repaint, faceswap} generate images with unprecedented photorealism, rendering existing tamper localization approaches~\cite{li2019localization,zhang2024editguard,xu2024fakeshield} increasingly inadequate.

Current tamper localization methods employ a post-hoc approach, analyzing content only after manipulation has occurred. Such a manner relies on specific forgery patterns learned during task-specific training~\cite{dong2022mvss, sun2023safl, ma2023iml, ramesh2022hierarchical, asnani2023malp}. However, such reliance undermines generalizability~\cite{zhang2024editguard} when encountering novel forgery contents absent from training data. As AI models advance toward generating increasingly diverse and novel content, these limitations will become more pronounced. To address this, we need \textbf{new thinking} beyond the current post-hoc strategy that relies on learned image tampering traces.

\begin{figure}[!t]
    \centering
    \includegraphics[width=1.0\linewidth]{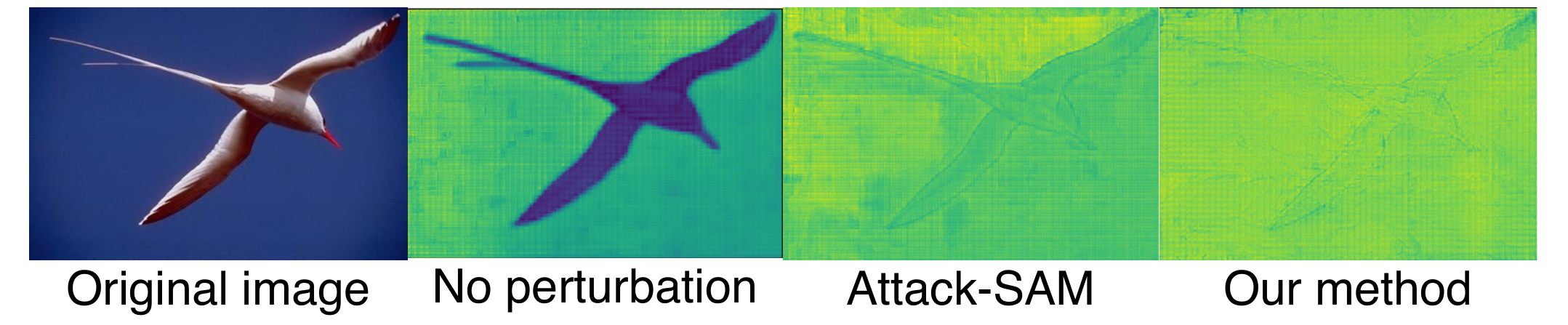}
   \vspace{-0.6cm}
    \caption{Predicted segmentation confidence maps from the segment anything model. Previous adversarial attack~\cite{zhang2023attack} could not fully disrupt the SAM's perception, especially in edges and texture areas.}
    \label{fig:teaser1}
   \vspace{-0.4cm}
\end{figure}

Recently, various powerful vision foundation models~\cite{kirillov2023segment} have shown remarkable generalizability across various tasks. A good way is to leverage their generalizability in tamper localization. However, acquiring adequate datasets to develop a foundation model for tampering detection remains prohibitively challenging. Moreover, even with sufficient data, training such a foundation model requires substantial computational resources~\cite{kwon2025safire}, which may be inaccessible to many photo owners. To address this, we introduce a proactive approach: instead of relying on post-hoc localization, photo owners could embed invisible protective layers into images before distribution. These layers enable an off-the-shelf foundation model (\eg, SAM) to automatically detect tampered regions without requiring task-specific tampering training.

\begin{figure*}[t]
  \centering
   \includegraphics[width=1.0\linewidth]{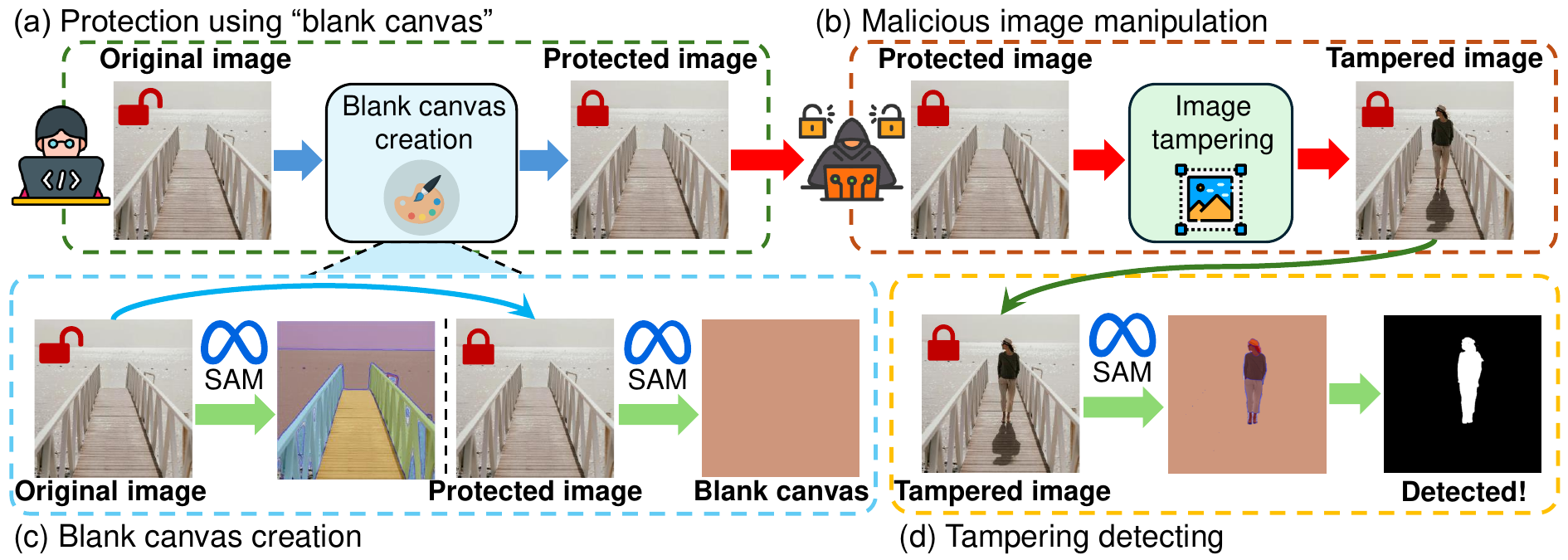}
   \vspace{-0.5cm}
   \caption{Overview of our tamper localization framework. (a) The protection process transforms an original image into a protected image through our blank canvas creation mechanism. (b) Illustration of potential malicious image manipulation on the protected image. (c) The blank canvas creation process demonstrates how the original image generates detailed SAM segmentation results while our protected image appears as a blank canvas to SAM~\cite{kirillov2023segment}. (d) The tampering detection phase shows SAM~\cite{kirillov2023segment} successfully identifying the tampered region when the protected image is maliciously manipulated.}
   \label{fig:teaser}
   \vspace{-0.2cm}
\end{figure*}

This paper proposes a novel ``blank-canvas'' mechanism to achieve this envisioned scenario, leveraging the insight that tampering is more discernible in a simplified context. As illustrated in \cref{fig:teaser}, tampering artefacts, often imperceptible in complex images due to complex textures, become prominently visible on a blank canvas. To exploit this,  we suggest transforming the complex image into a ``blank canvas'' in the perspective of vision foundation model. Then, once such a ``blank canvas'' is tampered, the vision foundation model can readily identify the tampered areas. In our setting, we propose using the Segment Anything Model (SAM)~\cite{kirillov2023segment}, the popular off-the-shelf foundation model with the required capability in perceiving visual content. Then, we incorporate adversarial perturbations to images. The adversarial perturbations are designed to mislead SAM's mechanism, directing its focus toward manipulated regions rather than authentic content.

However, SAM is a powerful model, and conventional adversarial attacks~\cite{madry2017towards,zhang2023attack}  cannot completely disrupt SAM perception (as further discussed in \cref{sec:abl}). Segmentation models like SAM are naturally designed to distinguish textural regions and structural boundaries within an image. However, as shown in \cref{fig:teaser1}, the high-frequency structural patterns inherent in complex images (\textit{e.g.}, edges, textures) retain perceptually salient contours. This challenge arises as SAM is inherently designed to identify different regions in an image based on the distinct features~\cite{kirillov2023segment}, which are often characterized by their frequency components in the spectral domain~\cite{zhou2024darksam,xia2023transferable}. 
Consequently, this leads to fragmented segmentation outputs instead of achieving a coherent suppression of structural elements. To effectively obscure SAM's perception in these high-frequency areas, we propose a Frequency-aware optimization that manipulates the image's frequency components, creating a more robust disguise against SAM's perception. Our strategy ensures that SAM perceives the protected image as a blank canvas, successfully enabling the transition from ``segment anything'' to ``segment nothing''. Malicious editing on the protected image could be identified as SAM perceives those areas of discrepancy. Our contributions can be summarized as:

\begin{itemize} 
\item We introduce a novel method for tamper localization that utilizes adversarial perturbations to hinder the segmentation capabilities of the SAM, thereby enhancing its ability to detect tampered regions. 
\item We present a new concept that edits made on a ``blank canvas'' are more conspicuous and easier to identify, thereby contributing to the broader field of image security. 
\item We propose a frequency-aware optimization strategy designed to deceive the segment anything model effectively. \end{itemize}

Extensive experimental results demonstrate the effectiveness of our solution for tamper localization.

\section{Related works}
\label{sec:related}
\subsection{Tamper localization}
The rapid advancements in AI-based image editing technology~\cite{rombach2022high, podell2023sdxl, zhang2023adding, lugmayr2022repaint, faceswap} have significantly benefited photographers and image editors by enabling unprecedented creative possibilities. However, these powerful capabilities also pose significant challenges for information security. Different tamper localization methods are proposed to address this challenge.
Existing approaches to image forensics predominantly concentrate on detecting predefined manipulation categories through passive analysis~\cite{li2018fast, zhu2018deep, wu2022robust, salloum2018image, islam2020doa, li2019localization,sun2024rethinking,sun2023safl,zhang2025omniguard}. Beyond these specialized solutions, broader-spectrum detection frameworks~\cite{ying2021image, kwon2021cat, chen2021image, wu2019mantra, li2018learning, ying2023learning, hu2023draw, ying2022rwn} attempt to identify tampering traces by analyzing inherent inconsistencies in forged images. Representative efforts include HiFi-Net~\cite{guo2023hierarchical}, which implements hierarchical feature extraction for both synthetic and edited content, and Trufor~\cite{guillaro2023trufor} that combines a transformer-enhanced fusion of noise patterns with spectral characteristics. While OSN~\cite{wu2022robust} enhances robustness against quality degradation through adaptive training strategies, SAFL-Net~\cite{sun2023safl} enforces manipulation-sensitive feature learning via auxiliary constraints. Despite these advancements, current passive detectors~\cite{dong2022mvss, sun2023safl, ma2023iml,ramesh2022hierarchical, asnani2023malp} frequently suffer from constrained generalizability and precision, typically excelling only on manipulation types encountered during training. Even proactive solutions like MaLP~\cite{asnani2023malp}, despite employing template-matching strategies, remain dependent on extensive forged samples and retain architecture-level coupling to specific tamper characteristics.
Recently, EditGuard~\cite{zhang2024editguard} proposed an active protection method that embeds information into images. Although effective in certain scenarios, this approach suffers from limited interpretability and practical constraints due to its reliance on non-intuitive steganographic content comparisons. To address these challenges, our method introduces visually detectable traces that enable transparent and human-verifiable tamper localization without requiring pre-registered steganographic references.

\subsection{Adversarial attacks on SAM}
\label{sec:adv_attacks_prelim}
As artificial intelligence systems become widely deployed, AI safety and assets are increasingly becoming a critical issue~\cite{huang2024geometrysticker,luo2023copyrnerf,luo2025nerf,luo2025imagesentinel,song2024geometry,song2024protecting,huang2025marksplatter,xu2025policy,li2025modeling,li2024variational,huang2024gaussianmarker,luo2025mantlemark}. Deep neural networks are widely known to be vulnerable to adversarial attacks CNN~\cite{szegedy2013intriguing,goodfellow2014explaining,kurakin2016adversarial} and vision transformer (ViT) ~\cite{dosovitskiy2020image,bhojanapalli2021understanding,mahmood2021robustness}. This vulnerability has inspired numerous works investigating the model's robustness under various adversarial attacks. Typical adversarial attacks have primarily focused on manipulating image-level label predictions in image classification tasks. For semantic segmentation, prior work~\cite {xie2017adversarial_adv_sem,arnab2018robustness_adv_sem,hendrik2017universal_adv_sem} studies adversarial attacks on traditional semantic segmentation models. Adversarial attacks on SAM differ from attacks on semantic segmentation models since the generated masks do not have semantic labels. Recent works~\cite{shen2024practical,xia2023transferable,zhou2024darksam,zhang2023attack} have extended adversarial attacks to the emerging Segment Anything Model (SAM)~\cite{kirillov2023segment} for prompt-based mask prediction.  SAM-attack~\cite{zhang2023attack} investigates the robustness of SAM against adversarial attacks. S-RA~\cite{shen2024practical} and Dark-SAM~\cite{zhou2024darksam} aim to build a unified, transfer-adversarial generation framework that combines a frequency-domain loss and an area-consistency loss. MUI-GRAT~\cite{xia2023transferable} investigates transferable attacks simultaneously compromising SAM and its downstream models, proposing a parameter-freezing strategy to enhance cross-model transferability. These studies reveal unique vulnerabilities in SAM's mask prediction paradigm, differing fundamentally from traditional semantic segmentation attacks due to SAM's class-agnostic segmentation nature.

\section{Preliminaries}
\label{sec:preliminaries}

\noindent \textbf{Segment anything model.} 
Segment Anything Model (SAM)~\cite{kirillov2023segment} is the first vision foundation model for image segmentation.  Unlike conventional semantic segmentation methods~\cite{minaee2021image_seg,haralick1985image_seg} that produce class-labeled masks, SAM outputs unlabeled segmentation results corresponding to prompt-specified regions. The fundamental data structure can be formally defined as a triplet $(x, \mathcal{P}, \mathcal{M})$, where $x \in \mathbb{R}^{H \times W \times 3}$ and $\mathcal{P}$ denotes input image and prompt vector (spatial coordinates or bounding boxes), $\mathcal{M} \in \{0,1\}^{H \times W}$ is target binary mask. For a given image $x$, multiple valid annotations can be generated through stochastic prompt sampling across the image plane, forming an augmented dataset: $\mathbb{D} = \{(x, \mathcal{P}_i, \mathcal{M}_i)\}_{i=1}^N$. The model's inference process can be expressed as:
\begin{equation}
\Phi = \text{SAM}(x, \mathcal{P}),
\label{eq:sam_forward}
\end{equation}
where $\Phi \in \mathbb{R}^{H \times W}$ denotes the confidence score matrix with equivalent spatial dimensions to the input image. The segmentation decision rule follows a threshold operation:
\begin{equation}
\mathcal{M}_{\text{pred}}[i,j] = 
\begin{cases} 
1 & \Phi[i,j] > 0,\\
0 & \text{otherwise}.
\end{cases}
\end{equation}

The resultant binary mask $\mathcal{M}_{\text{pred}}$ preserves the original image resolution through its $H \times W$ dimensional structure. Notably, this architecture enables dynamic mask generation conditioned on various prompt types (points, boxes, or text) while maintaining spatial coherence in the output.

\noindent \textbf{Our scenario.} We aim to force SAM to perceive the protected image as a \textit{blank canvas}. Then, any subsequent image manipulation could lead SAM to identify the tampered regions as anomalous segmentation responses. This analogy is to detecting chalk marks on a blank canvas: 1) Before tampering, SAM produces no masks; 2) Tampering acts like chalk strokes on a blank canvas, making altered pixels ``noticeable'' in SAM's output.

During usage, image owners can implement our method and transform their images into a ``blank canvas'' from the SAM's perspective before distribution. These images appear unchanged, as the added perturbations are invisible and do not significantly affect the original image quality. However, if malicious users attempt to alter these protected images, the modifications would be noticeable to the SAM models, such as those made on a blank canvas. Users and third-party investors can directly access the anomalous areas identified by a naive SAM model.

\section{Proposed method}
\label{sec:method}
This section presents our training-free approach for adapting SAM as a foundation model for tamper localization. Our approach contains two stages: 1) \textit{Blank canvas creation}, which involves strategically suppressing SAM's segmentation capabilities across the images, thereby creating a perceived ``blank canvas'' from SAM's view. 2) \textit{Tamper localization}, during which, if any tampering occurs, the manipulated areas become noticeable from SAM's perspective, allowing it to identify tampered regions.

\begin{figure*}[t]
\centering
\includegraphics[width=0.90\textwidth]{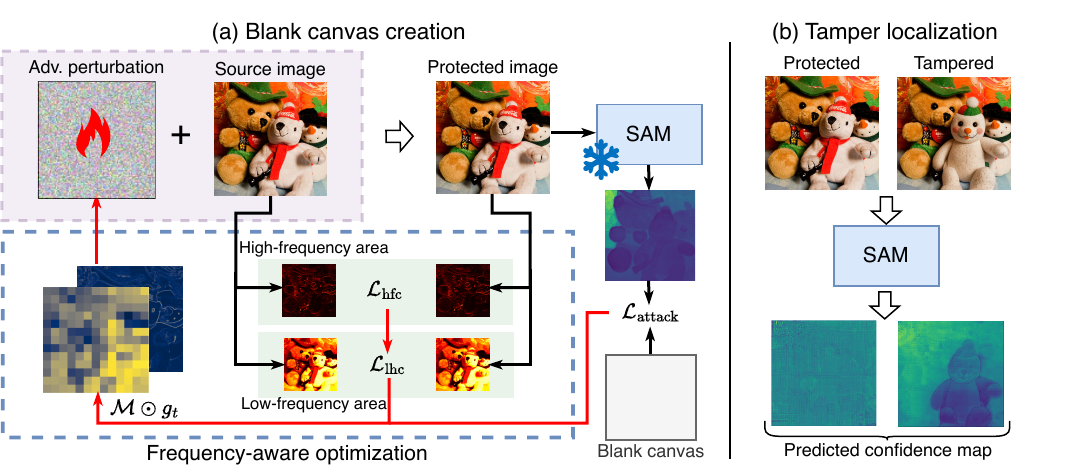}
\vspace{-0.45cm}
\caption{Overall of our method.  We enable the source image to be a ``blank canvas'' from the perspective of SAM. Frequency-aware optimization  is proposed to disrupt the high-frequency areas to deceive the SAM model fully. After the image is protected, the tampered locations become noticeable to SAM.}
\label{fig:pipeline}
\vspace{-0.45cm}
\end{figure*}

\subsection{Blank canvas creation}
Instead of identifying tampered areas through learned artifacts passively, we proactively transform the image into a ``blank canvas'', an intentional special state where SAM's segmentation confidence is uniformly distributed. Any further manipulations inevitably introduce abnormal deviations, which SAM recognizes as anomalous segmentation targets, effectively identifying the tampered area.

Specifically, we aim to force SAM to perceive the protected image as a blank canvas, \ie, a state where all segmentation confidence scores converge to a certain value \(c\) (\ie, \(\Phi[i,j] \approx c,\ \forall i,j\)). As described in Equation~\ref{eq:sam_forward}, a pixel \( x_{ij} \) is classified as masked if the predicted value \( y_{ij} \) is positive. The value \( y_{ij} \in \Theta \) is derived from SAM and indicates the confidence that the position \( \{i,j\} \) should be masked. Consequently, if the predicted values \( y \) across the images converge towards a typical constant, SAM interprets this image as a blank canvas. Specifically, our framework aims to identify such a perturbation \( \delta \):
\begin{equation}
\Phi' = \text{SAM}(x_{\text{clear}} + \delta, \mathcal{P}),
\label{eq:adv_aim}
\end{equation}
where \( x_{\text{clear}} \) represents the images requiring protection and \( \mathcal{P} \) denotes the point prompt for SAM. The perturbation \( \delta \) is designed to ensure that \( \Phi' \) (the output of SAM) approaches a constant \( C \). If every element in the matrix \( \Phi' \) converges to this common constant, the entire image is perceived as a ``blank canvas'' from SAM's perspective.

To guarantee that SAM's output approaches the constant \( C \), we employ the Mean Squared Error (MSE) loss, which is a natural fit for this optimization task. As articulated in Equation~\ref{eq:mse_loss}, the predicted value \( \text{SAM}(\text{prompt}, x_{\text{clean}} + \delta) \) is optimized to be close to a target threshold after the attack. Thus, we aim for the predicted values to equal the constant \( C \), leading to the loss function defined as:
\begin{equation}
    \mathcal{L}_\text{attack} =  \texttt{MSE}(\Phi, C),
    \label{eq:mse_loss}
\end{equation}
where \( \Phi \) represents the output of SAM. The MSE loss in Equation~\ref{eq:mse_loss} facilitates the prediction of negative values, aligning with our expectations for the perturbation \( \delta \). This loss term ensures that the predicted results remain close to the constant \( C \), effectively rendering the images as a unified whole from SAM’s perspective.

We can effectively obscure critical information that the model relies on for accurate segmentation by compelling SAM to perceive the protected images as blank canvases. However, we notice that the naive adversarial attack, \ie, above $\mathcal{L}_\text{attack}$, is insufficient to deceive SAM, especially in high-frequency edges (further demonstrated in ablation study). To address this issue, we further introduce a \textbf{\textit{Frequency-aware optimization}}, which is designed to enhance adversarial perturbations through the frequency domain with adaptive spectral optimization. It comprises three synergistic components, designed to force adversarial perturbations that disrupt high-frequency areas while preserving the fidelity of low-frequency regions.

\subsubsection{\textbf{Wavelet-domain frequency decomposition.}}
Firstly, we target the high-frequency perturbations that disrupt SAM's edge detection capabilities. We decompose the source image \( x \) and perturbed image \( \tilde{x} \) into the frequency domain using the discrete wavelet transform (DWT) with the Daubechies-8 basis. This allows us to extract critical high-frequency components for disrupting SAM's edge perception. 
The loss can be computed as follows:
\begin{equation}
\mathcal{L}_{\text{hfc}} = \sum_{k=1}^K \|\underbrace{\mathcal{W}_k(\tilde{x}) \odot M_{\text{edge}}}_{\text{perturbed edges}} - \underbrace{\mathcal{W}_k(x) \odot M_{\text{edge}}}_{\text{original edges}}\|_F^2.
\label{eq:wavelet_loss}
\end{equation}
Here, \( \mathcal{W}_k \) extracts wavelet coefficients at level \( k \), and \( \odot \) denotes the Hadamard product. The high-frequency perturbations are constrained by a Canny edge mask \( M_{\text{edge}} \). This strategy aims to disrupt the high-frequency areas of the image to tame SAM fully for accurate tamper localization.

\subsubsection{\textbf{Structural preservation constraint.}} While we disrupt high-frequency information, the overall visual integrity of the image remains intact. This is crucial for avoiding detection by human observers or automated systems. We incorporate a structural preservation constraint that safeguards the low-frequency components using an adaptive Structural Similarity Index (SSIM) to maintain visual naturalness and prevent excessive distortion.  The SSIM-based loss is defined as:
\begin{equation}
\mathcal{L}_{\text{lfc}} = \text{SSIM}(\phi_m ,\tilde{\phi}_m ),
\label{eq:ssim}
\end{equation}
where \( \phi_m  = \mathcal{L}_m^T x_{\text{lfc}}^{(m)} \mathcal{L}_m \), representing reconstructs the low frequency components at scale \( m \). \( \tilde\phi_m \).
This component ensures that the low-frequency characteristics of the image are preserved, maintaining its natural appearance while still achieving the desired adversarial effect.

\subsubsection{\textbf{Adaptive spectral optimization.}} Finally, we introduce frequency-aware adversarial optimization with momentum, as detailed in \cref{alg:fgsm}. The motivation for this optimization approach is to derive the optimal perturbation \( \delta^* \) that maximizes the effectiveness of the attack while adhering to constraints on perturbation magnitude. The optimization problem is formulated as:
\begin{equation}
\delta^* = \arg\max_{\|\delta\|_\infty \leq \epsilon} \underbrace{\mathbb{E}_{p\sim\mathcal{P}}[\|SAM(x+\delta,p)\|_1 ]}_{\mathcal{L}_{\text{attack}}} + \underbrace{\lambda {\mathcal{L}_{\text{lfc}}} -\beta \mathcal{L}_\text{hfc}}_{\mathcal{L}_\text{stealth}}.
\label{eq:adv_obj_updated}
\end{equation}
In this context, \( \mathcal{L}_{\text{attack}} \) combines mask prediction suppression and edge disruption, defined as follows:
\begin{equation}
\mathcal{L}_{\text{all}} = \underbrace{\frac{1}{N}\sum_{i=1}^N \|SAM(x+\delta,p_i)\|_2^2}_{\text{MSE suppression}} + \mathcal{L}_{\text{stealth}}.
\end{equation}

\begin{algorithm}[t]
\caption{Adaptive spectral optimization}
\label{alg:fgsm}
\begin{algorithmic}[1]
\Require  $\mathcal{L}_{\text{attack}}$, $\mathcal{L}_{\text{stealth}}$,  Fourier transform $\mathcal{F}$, iterations $T$, step size $\alpha_0$, perturbation bound $\epsilon$, momentum decay $\mu \in [0,1)$
\Ensure Adversarial perturbation $\delta$
\State Initialize $\delta_0 \sim \mathcal{U}(-\epsilon, \epsilon)$
\State Initialize momentum: $m_0 = 0$
\For{t = 0 to T-1}
  \State  $g_t = \nabla_\delta (\mathcal{L}_{\text{attack}} + \mathcal{L}_{\text{stealth}})$ \Comment{Compute gradient}
  \State $\hat{g}_t = \mathcal{F}^{-1}(\mathcal{F}(g_t) \odot \mathcal{M})$ \Comment{Spectral projection}
  \State  $m_{t+1} = \mu m_t + \frac{\hat{g}_t}{\|\hat{g}_t\|_1}$  \Comment{Update momentum}
  \State $\alpha_t = \alpha_0(1 - e^{-5t/T})$  \Comment{Adaptive step}
  \State $\delta_{t+1} = \text{Clip}_\epsilon[\delta_t + \alpha_t \cdot \text{sign}(m_{t+1})]$   \Comment{Update $\delta$}
\EndFor
\end{algorithmic}
\end{algorithm} 

The loss item $\mathcal{L}_\text{attack}$ is used to optimize the adversarial perturbation. We adaptively optimize the gradient based on the perturbation energy to maximize its impact on the high-frequency contents and minimize its impact on the low-frequency area. Specifically, a spectral projection mask \( \mathcal{M} \) is designed to focus perturbation energy on high-frequency bands, defined as:
\begin{equation}
\mathcal{M}(u,v) = 
\begin{cases}
1, & \sqrt{u^2 + v^2} \geq f_{\text{cutoff}}, \\
0, & \text{otherwise}.
\end{cases}
\end{equation}
where \( f_{\text{cutoff}} \) acts as the cutoff frequency controlling the locality of the perturbation. \( u \) and \( v \) denote the frequency components in the frequency domain of the source image. We then use this map to control the adversarial optimization process via \cref{alg:fgsm}. 

\subsection{Tamper localization}

Following the blank canvas initialization, the adversarial image \( \tilde{x} = x + \delta \) must satisfy the condition:
\begin{align}
\Phi_{\text{adv}}[i,j] &= SAM(\tilde{x})[i,j] \notag \approx C, \\
&\quad \forall i,j \in \{1,...,H\}\times\{1,...,W\}.
\label{eq:blank_condition}
\end{align}
When tampering occurs on the blinded image \( \tilde{x} \), the alteration \( \Delta x \) generates localized perturbation breaches, which can be captured as:

\begin{equation}
\mathcal{M}_{\text{tamper}} = \mathbb{I}\left(\|SAM(\tilde{x} + \Delta x)\|_2 > \tau_{\text{detect}}\right).
\label{eq:detection}
\end{equation}

In this equation, \( \tau_{\text{detect}} \) is adaptively determined using Otsu's method~\cite{otsu1975threshold_otsu}. The tampered areas $\mathcal{M}_\text{tamper}$ are determined through evaluating the noticeable areas in the confidence map predicted by SAM.

\subsection{Implementation details}

To optimize the adversarial perturbations, we adopt the projected gradient descent (PGD) attack framework~\cite{madry2017towards}. In alignment with prior research focused on attacking vision models in a white-box setting, we establish the maximum allowable perturbation magnitude at \( 16/255 \), with a step size of \( 2/255 \) for PGD. The optimized target constant \( C \) is set to \( 15 \), as the predicted values in the background (non-mask) region typically hover around this value. For our experiments, we utilize the vanilla version of SAM~\cite{kirillov2023segment} with ViT-H~\cite{dosovitskiy2020image} backbone and employ a single point prompt located at \( (0,0) \). This setup enables the prediction of a single mask in proximity to that point. Given that each mask is generated based on an input prompt, the attack is deemed successful if SAM fails to predict the original mask corresponding to that prompt accurately. Specifically, the attack succeeds in the mask removal task if \( \text{Mask}_{\text{adv}} \) is empty or if its area is at least as small as that of \( \text{Mask}_{\text{clean}} \). To maintain generality, the \( (\text{prompt}, \text{Mask}) \) pair selected for the attack is randomly chosen from the image.

\begin{figure*}[!t]
    \centering
    \includegraphics[width=0.85\linewidth]{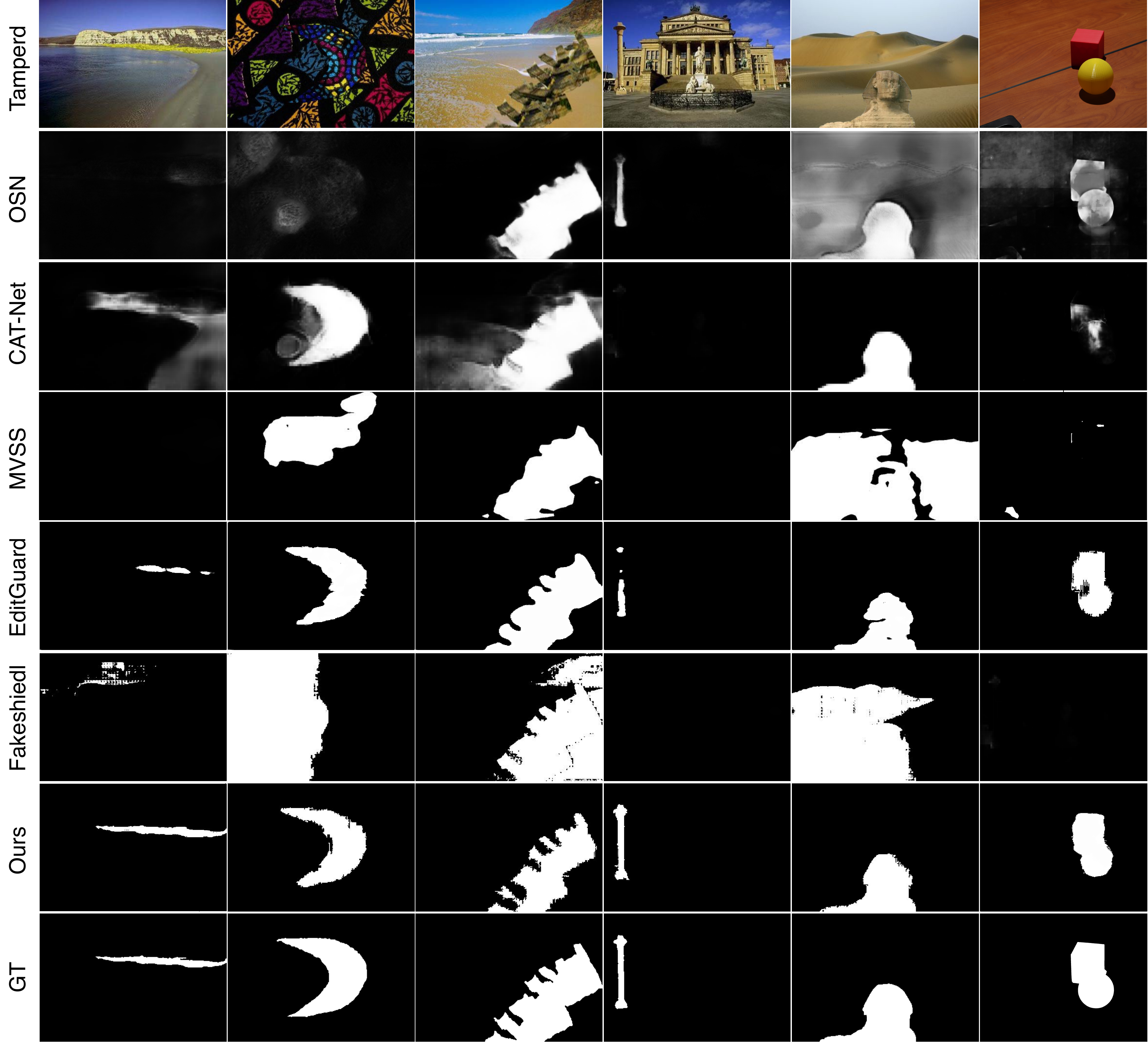}
    \vspace{-0.3cm}
	\caption{Visual results of localized tampering areas. Our method achieves superior performance compared to previous passive tamper localization methods, including OSN~\cite{wu2022robust}, CAT-Net~\cite{kwon2021cat}, MVSS~\cite{dong2022mvss}, and FakeShield~\cite {xu2024fakeshield}. We also achieve comparable performance compared to protective methods EditGuard~\cite{zhang2024editguard} across each scene.}
    \vspace{-0.35cm}
    \label{fig:classical}
\end{figure*}

\section{Experiments}
\noindent \textbf{Benchmarks.}  To comprehensively evaluate our method, we conduct experiments across four tamper localization dataset~\cite{dong2013casia, wen2016coverage, guan2019mfc, hsu2006detecting}, comparing against eight state-of-the-art methods including SPAN~\cite{hu2020span}, ManTraNet~\cite{wu2019mantra}, OSN~\cite{wu2022robust}, HiFi-Net~\cite{guo2023hierarchical}, PSCC-Net~\cite{liu2022pscc}, CAT-Net~\cite{kwon2021cat}, MVSS-Net~\cite{dong2022mvss}, and FakeShield~\cite{xu2024fakeshield}. To simulate real-world forgery scenarios, we further evaluate on the AIGC-based editing dataset~\cite{zhang2024editguard}. Editing methods, including SD Inpaint~\cite{rombach2022high}, Controlnet~\cite{zhang2023adding}, SDXL~\cite{podell2023sdxl} are used to manipulate images with the prompt to be ``None''. For fair comparison with EditGuard~\cite{zhang2024editguard} - the current state-of-the-art in proactive protection, the images are protected in advance before tampering. Following~\cite{dong2022mvss, guillaro2023trufor, liu2022pscc,zhang2024editguard}, F1-score, IoU are used to evaluate the quality of tamper localization. Higher F1-score and IoU indicate better localization performance.

\begin{table*}[t]
\centering
\renewcommand{\arraystretch}{1.0}
\resizebox{0.85\linewidth}{!}{
\begin{tabular}{ccccccccccccc}
\toprule
\multirow{2}{*}{\makecell{\centering Method}} & \multicolumn{2}{c}{CASIA1+} & \multicolumn{2}{c}{IMD2020} & \multicolumn{2}{c}{Columbia} & \multicolumn{2}{c}{NIST} & \multicolumn{2}{c}{DSO} & \multicolumn{2}{c}{Korus}  \\
& IoU & F1 & IoU & F1 & IoU & F1 & IoU & F1 & IoU & F1 & IoU & F1 \\
\midrule
SPAN~\cite{hu2020span} & 0.11 & 0.14 & 0.09 & 0.14 & 0.14 & 0.20 & 0.16 & 0.21 & 0.14 & 0.24 & 0.06 & 0.10  \\
ManTraNet~\cite{wu2019mantra} & 0.09 & 0.13 & 0.10 & 0.16 & 0.04 & 0.07 & 0.14 & 0.20 & 0.08 & 0.13 & 0.02 & 0.05 \\
OSN~\cite{wu2022robust} & 0.47 & 0.51 & 0.38 & 0.47 & 0.58 & 0.69 & 0.25 & 0.33 & 0.32 & 0.45 & 0.14 & 0.19  \\
HiFi-Net~\cite{guo2023hierarchical} & 0.13 & 0.18 & 0.09 & 0.14 & 0.06 & 0.11 & 0.09 & 0.13 & 0.18 & 0.29 & 0.01 & 0.02  \\
PSCC-Net~\cite{liu2022pscc} & 0.36 & 0.46 & 0.22 & 0.32 & 0.64 & 0.74 & 0.18 & 0.26 & 0.22 & 0.33 & 0.15 & 0.22  \\
CAT-Net~\cite{kwon2021cat} & 0.44  & 0.51  & 0.14  & 0.19  & 0.08  & 0.13  & 0.14  & 0.19  & 0.06  & 0.10  & 0.04  & 0.06  \\
MVSS-Net~\cite{dong2022mvss} & 0.40  & 0.48  & 0.23  & 0.31  & 0.48  & 0.61  & 0.24  & 0.29  & 0.23  & 0.34  & 0.12  & 0.17 \\
FakeShield~\cite{xu2024fakeshield} & 0.56 & \underline{0.62} & 0.52 & 0.58 & 0.68 & 0.76 & 0.34 & 0.39 & 0.50 & 0.54 & \underline{0.22} & 0.26  \\
EditGuard~\cite{zhang2024editguard} & \underline{0.60} & \textbf{0.67} & \underline{0.55} & \underline{0.62} & \underline{0.70} & 0.78 & \textbf{0.35} & 0.40 & \underline{0.52} & 0.56 & \underline{0.22} & \underline{0.28}  \\
Ours & \textbf{0.62} & \textbf{0.67} & \textbf{0.58} & \textbf{0.66} & \textbf{0.74} & \textbf{0.81} & \underline{0.31} & \textbf{0.45} & \textbf{0.55} & \textbf{0.60} & \textbf{0.27} & \textbf{0.31}  \\
\bottomrule
\end{tabular}}
\vspace{-0.3cm}
\caption{Comparison with other tamper localization methods on the classical tamper localization dataset. }
\label{tab:classical}
\end{table*}

\begin{table*}[!t]
\centering
\resizebox{0.85\linewidth}{!}{%
\begin{tabular}{ccccccccccc}
\toprule[0.8pt]
\multicolumn{1}{c}{\multirow{2}{*}{Method}} & \multicolumn{2}{c}{SD Inpaint} & \multicolumn{2}{c}{ControlNet} & \multicolumn{2}{c}{SDXL} & \multicolumn{2}{c}{RePaint} & \multicolumn{2}{c}{Lama} \\ 
 & F1 & IoU  & F1 & IoU  & F1 & IoU  & F1 & IoU  & F1 & IoU   \\ 
\midrule
MVSS-Net~\cite{dong2022mvss}  & 0.178 & 0.103  & 0.178 & 0.103  & 0.037 & 0.028  & 0.104 & 0.082  & 0.024 & 0.022    \\
PSCC-Net~\cite{liu2022pscc} & 0.166  & 0.112  & 0.177  & 0.116  & 0.189  & 0.115  & 0.140  & 0.109  & 0.132  & 0.104   \\
HiFi-Net~\cite{guo2023hierarchical} & 0.547  & 0.128  & 0.542  & 0.123  & 0.828 & 0.261   & 0.681   & 0.339    & 0.483  & 0.029   \\
$\text{MVSS-Net} ^{\dagger} $~\cite{dong2022mvss}   & 0.694   & 0.575    & 0.678   & 0.558   & 0.482  & 0.359   & 0.185 & 0.111  & 0.393  & 0.275    \\
    EditGuard~\cite{zhang2024editguard} & \underline{0.966} & \underline{0.936} & \underline{0.968} & \textbf{0.940} & \underline{0.965} & \underline{0.936} & \textbf{0.967} & \underline{0.938} & \textbf{0.965} & \underline{0.934}  \\ 
Ours & \textbf{0.972} & \textbf{0.958} & \textbf{0.973} & \underline{0.938} & \textbf{0.970} & \textbf{0.958} & \underline{0.961} & \textbf{0.957} & \underline{0.954} & \textbf{0.951} \\
\bottomrule[0.8pt]
\end{tabular}}
\vspace{-0.20cm}
\caption{Comparison with other competitive tamper localisation methods under different AIGC-based editing methods. Note that $\dagger$ denotes the network fine-tuned with the AIGC-edited dataset~\cite{zhang2024editguard}.}
\label{tab:aigc}
\end{table*}

\begin{figure*}[!t]
	\centering
    \includegraphics[width=0.90\linewidth]{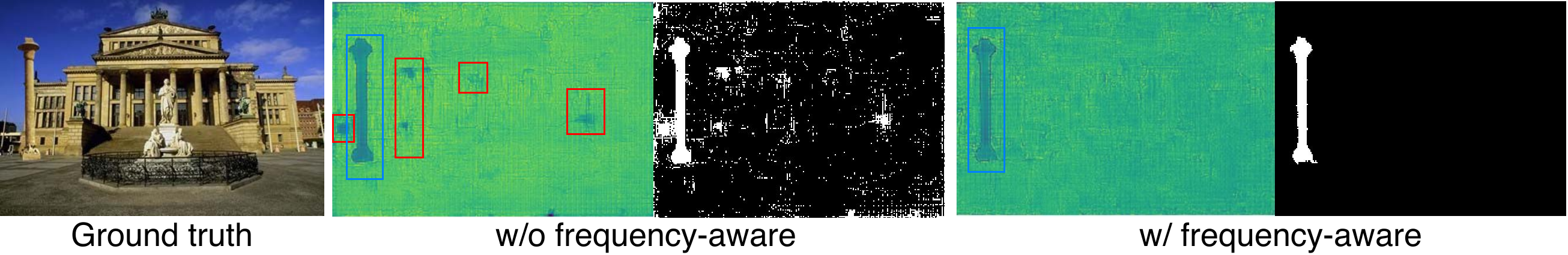}
    \vspace{-0.20cm}
	\caption{Ablation study on the proposed frequency-aware optimization. False positive results could occur in high-frequency areas as they are more likely to be noticed by SAM. The red/blue box indicates the false/true positives of localized regions.}
    \label{fig:visual1}
\end{figure*}

\begin{table}[!t]
    \centering
    \resizebox{0.90\linewidth}{!}{
    \begin{tabular}{c|c|ccc|ccc}
    \toprule[1.5pt] 
    Case & Degradation$^*$ & $\mathcal{L}_\text{mse}$ & $\mathcal{L}_{\text{stealth}}$ & \texttt{Ada.} & F1 & IoU \\ 
    \hline 
    (a) & Clean & & & &  0.352 & 0.378 \\
    (b) & Clean & $\checkmark$ & $\checkmark$  & & 0.934 & 0.928 \\
    (c) & Clean & $\checkmark$ &  &  $\checkmark$ & 0.931 & 0.921 \\
    \hline 
    \multirow{2}{*}{Ours} 
    & Clean & $\checkmark$ & $\checkmark$ & $\checkmark$ & 0.964 & 0.955 \\ 
    & Random & $\checkmark$ & $\checkmark$ & $\checkmark$ & 0.945 & 0.931 \\ \bottomrule[1.5pt] 
    \end{tabular}}
    \vspace{-0.2cm}
    \captionof{table}{Our frequency-aware optimization, \ie, combing $\mathcal{L}_{\text{stealth}}$ with \texttt{Ada.}, achieves the best results as it could fully disrupt the perception feature of SAM. Case (a) denotes that the images are not being protected with any adversarial perturbations. } 
    \label{tab:ablation_study}
    \vspace{-0.1cm}
\end{table}

\subsection{Experimental results}

\noindent \textbf{Results on classical benchmarks.}
For a comprehensive comparison with existing tamper localization methods, we conduct evaluations using classical benchmarks~\cite{dong2013casia, wen2016coverage, guan2019mfc, hsu2006detecting}, as summarized in \cref{tab:classical}. Our method demonstrates competitive performance against the previous state-of-the-art method EditGuard~\cite{zhang2024editguard}. In several cases, we achieve marginally better localization accuracy, surpassing it by small margins in F1-score across various datasets. respectively. This indicates the effectiveness of our proactive localization mechanism, which does not require any labeled data or tampered samples. As depicted in \cref{fig:classical}, our method effectively identifies pixel-level tampered areas, similar to EditGuard, while other methods tend to produce only rough outlines or show effectiveness in limited scenarios. This highlights the robustness and reliability of our approach in the realm of tamper localization.

\noindent \textbf{Results on AIGC-based editing methods.} We conduct experiments on various AIGC-based image editing methods using the AGE-Set dataset~\cite{zhang2024editguard}, following the evaluation framework established by EditGuard~\cite{zhang2024editguard}. The comparison of our method with several state-of-the-art tamper localization techniques is presented in Tab.~\ref{tab:aigc}. We observe that the F1-scores of existing passive forensic methods are generally below 0.7 when applied to AIGC-edited images. Notably, even after fine-tuning MVSS-Net on the constructed AIGC-edited dataset, the performance of $\text{MVSS-Net}^{\dagger}$ remains inadequate, exhibiting significant limitations in handling various tampering techniques. In contrast, our method consistently achieves F1 Scores and AUCs exceeding 95\%, while maintaining an IoU of around 90\% across different tampering types. Our approach effectively captures subtle tampering traces introduced by AIGC-based editing methods, whereas other competing methods struggle to produce meaningful results.

\noindent \textbf{Ablation study.}
\label{sec:abl}
To evaluate the contribution of each component in our proposed method, we conduct ablation studies on our frequency-aware perturbation (\cref{tab:ablation_study}), visualized in \cref{fig:visual1}, as well as the mean squared error loss ($\mathcal{L}_\text{mse}$) and stealth loss ($\mathcal{L}_{\text{stealth}}$). We observe that omitting frequency-aware optimization results in substantial performance degradation, reducing effectiveness to the level of unprotected images. In contrast, incorporating $\mathcal{L}_\text{mse}$ markedly improves results, with further gains from adding $\mathcal{L}_{\text{stealth}}$, underscoring their essential roles in disrupting perception features and boosting tampering-detection accuracy. Beyond these core components, supplemental experiments affirm robustness under both clean conditions and random degradations, while additional validation across diverse SAM variants demonstrates consistent effectiveness (please see supplement).
\section{Conclusion}
This paper introduces a novel proactive framework for tamper localization by transforming images into machine-interpretable ``blank canvases'' via frequency-optimized adversarial perturbations applied to vision foundation models. By suppressing the model's perception of original content while amplifying its sensitivity to synthetic alterations, our method addresses the limitations of conventional passive forensic approaches that rely on artifact detection or tampered training data. Our method shifts from post-hoc analysis to proactive protection and demonstrates strong generalizability across diverse manipulation scenarios, offering a scalable solution for real-world image authentication.

\clearpage

\section*{Acknowledgments}
This work was carried out at the Renjie Group, Hong Kong Baptist University. Renjie Group is supported by the National Natural Science Foundation of China under Grant No. 62302415, Guangdong Basic and Applied Basic Research Foundation under Grant No.  2024A1515012822, and Research Grant Council (RGC) of Hong Kong SAR, under a GRF Grant 12203124 and an ECS Grant 22201125.

\bibstyle{aaai2026}
\bibliography{reference}


\end{document}